\documentclass[conference]{IEEEtran}
\IEEEoverridecommandlockouts
\usepackage{cite}
\usepackage{amsmath,amssymb,amsfonts}
\usepackage{algorithmic}
\usepackage{graphicx}
\usepackage{textcomp}  
\usepackage{xcolor}
\usepackage{adjustbox}
\usepackage{xspace}
\usepackage{multirow}
\usepackage{authblk}
\usepackage[pagebackref=true,breaklinks=true,colorlinks,bookmarks=false]{hyperref}

\makeatletter
\DeclareRobustCommand\onedot{\futurelet\@let@token\@onedot}
\def\@onedot{\ifx\@let@token.\else.\null\fi\xspace}

\def\etal{\emph{et al}\onedot}
\makeatother

\def\BibTeX{{\rm B\kern-.05em{\sc i\kern-.025em b}\kern-.08em
    T\kern-.1667em\lower.7ex\hbox{E}\kern-.125emX}}
\begin{document}

\title{Implicit Multi-Spectral Transformer: An Lightweight and Effective Visible to Infrared Image Translation Model}

\author[1]{Yijia Chen}
\author[1\IEEEauthorrefmark{1}]{Pinghua Chen\thanks{\IEEEauthorrefmark{1} Corresponding author}}
\author[1]{Xiangxin Zhou}
\author[2]{Yingtie Lei}
\author[3]{Ziyang Zhou}
\author[3]{Mingxian Li}
\affil[1]{Guangdong University of Technology, Guangzhou, China}
\affil[2]{University of Macau, Macau, Macao}
\affil[3]{Huizhou University, Huizhou, China}

\maketitle

\begin{abstract}
In the field of computer vision, visible light images often exhibit low contrast in low-light conditions, presenting a significant challenge. While infrared imagery provides a potential solution, its utilization entails high costs and practical limitations. Recent advancements in deep learning, particularly the deployment of Generative Adversarial Networks (GANs), have facilitated the transformation of visible light images to infrared images. However, these methods often experience unstable training phases and may produce suboptimal outputs. To address these issues, we propose a novel end-to-end Transformer-based model that efficiently converts visible light images into high-fidelity infrared images. 
Initially, the Texture Mapping Module and Color Perception Adapter collaborate to extract texture and color features from the visible light image. The Dynamic Fusion Aggregation Module subsequently integrates these features. Finally, the transformation into an infrared image is refined through the synergistic action of the Color Perception Adapter and the Enhanced Perception Attention mechanism.
Comprehensive benchmarking experiments confirm that our model outperforms existing methods, producing infrared images of markedly superior quality, both qualitatively and quantitatively. Furthermore, the proposed model enables more effective downstream applications for infrared images than other methods.
\end{abstract}


\section{Introduction}
In the field of computer vision, visible light images (VIS) are commonly used as training data to train models. During the imaging process in visible light, there are certain situations where acquired images suffer from low contrast or certain objects demonstrate diminished reflectance in the visible spectrum. Infrared imagery (IR) can mitigate these issues, offering enhanced contrast and detectability of objects regardless of their visible light reflectivity. For instance, visible light images obtained under low light conditions are susceptible to issues such as low contrast, blurred edges, and missing details. Domains that have high demand for edge information—including pedestrian detection~\cite{jia2021llvip,liu2019ptb}, autonomous driving~\cite{Hwang2015MultispectralPD,ha2017mfnet}, robotic navigation~\cite{Zhang_CVPR22_VTUAV,wang2022applications} and precision agriculture~\cite{Chiu_2020_CVPR,chen2020deep}—may see substantial improvements from the superior edge definition offered by infrared imaging. Such enhancements are particularly crucial for safety-critical applications like autonomous driving. Nevertheless, the cost of infrared imaging equipment remains significant in comparison to widely accessible visible light cameras.

\begin{figure}[t]
  \centering
  % 第一行的第一个图像
  \begin{minipage}[c]{1.0\linewidth}
    \begin{minipage}[b]{0.32\linewidth}
        \centering
        \centerline{\includegraphics[width=\linewidth]{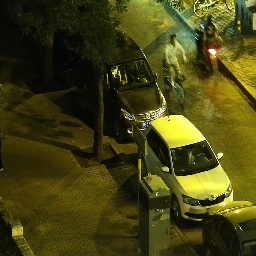}}
        \centerline{(a) Input}\medskip
    \end{minipage}
    \begin{minipage}[b]{0.32\linewidth}
        \centering
        \centerline{\includegraphics[width=\linewidth]{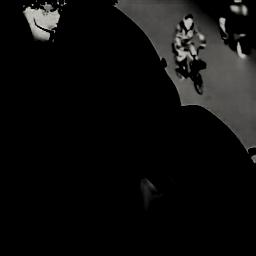}}
        \centerline{(b) CycleGAN}\medskip
    \end{minipage}
    \begin{minipage}[b]{0.32\linewidth}
        \centering
        \centerline{\includegraphics[width=\linewidth]{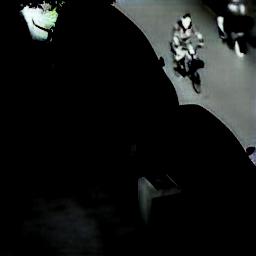}}
        \centerline{(c) MUNIT}\medskip
    \end{minipage}
  \end{minipage}
  \begin{minipage}[c]{1.0\linewidth}
    \begin{minipage}[b]{0.32\linewidth}
        \centering
        \centerline{\includegraphics[width=\linewidth]{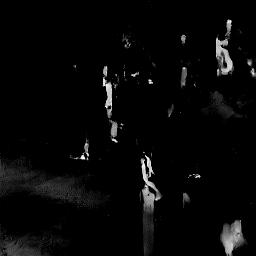}}
        \centerline{(d) BCI}\medskip
    \end{minipage}
    \begin{minipage}[b]{0.32\linewidth}
        \centering
        \centerline{\includegraphics[width=\linewidth]{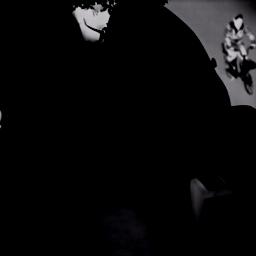}}
        \centerline{(e) ClawGAN}\medskip
    \end{minipage}
    \begin{minipage}[b]{0.32\linewidth}
        \centering
        \centerline{\includegraphics[width=\linewidth]{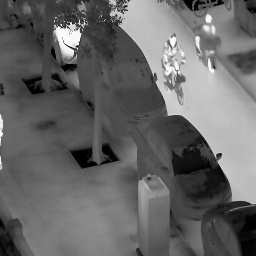}}
        \centerline{(f) Target}\medskip
    \end{minipage}
  \end{minipage}
  \caption{
   The comparison of various image translation techniques for Visible-to-Infrared image translation is illustrated as follows: (a) presents the original visible light image; (b) - (e) depict the translated images as produced by CycleGAN, MUNIT, BCI, and ClawGAN, respectively; (f) showcases the reference thermal image used as the translation target.
  } 
\label{fig:teaser}
\end{figure}

As a result of the adversarial training between generative and discriminative components, GANs~\cite{NIPS2014_5ca3e9b1} demonstrate superior performance in image generation tasks compared to Convolutional Neural Networks (CNNs). Applications utilizing GANs for image translation, such as segmantation~\cite{liu2023explicit,liu2024depth}, semantic-to-vision conversion~\cite{dong2017semantic,gong2023generative,cherian2019sem,li2023cee,li2022monocular,zhou2024qean,zuo2023diffgan}, super-resolution~\cite{wang2018esrgan,jiang2019edge}, image synthesis~\cite{brock2018large,zhan2019spatial,liu2023coordfill}, and style transfer~\cite{azadi2018multi,kwon2022clipstyler}, have reported noteworthy successes. Nonetheless, when GANs are employed for VIS-IR image translation, outcomes may be suboptimal. For example, although prevalent GAN models have the capability to effectively extract contrast features from infrared images, they often tend to misjudge other luminosity-correlated features, resulting in fabricated situations and irregular visual artifacts in areas where no objects exist as shown in Figure~\ref{fig:teaser} (b) - (e). In the analysis of the image translation techniques detailed in parts (b), (c), and (e) of the document, a common issue observed is the significant coverage of feature information by black pixels. This overwriting of information results in a loss of critical details that are essential for accurate image representation in the infrared domain. As for (d) BCI, the methodology appears to introduce a different kind of artifact—irrelevant ghosting or duplications of objects not present in the original scene. Additionally, diffusion models are not ideal for this task due to their consumption of extensive resources and time~\cite{huang2023mr,10365931,Chen2024-dg}.

To address these issues, we propose IRFormer, a novel, end-to-end Transformer-based model specifically designed for VIS-IR image translation. This model harnesses a Color Perception Adapter (CPA) to extract RGB information from visible light images, as well as a Enhanced Feature Mapping Module (EFM) to capture intricate textural details. Subsequently, these features are combined via the Dynamic Fusion Aggregation Module (DFA), which effectively translates the features into a latent representation that mediates between the VIS and IR domains. The CPA plays a crucial role in mapping color attributes onto IR images, while an Enhanced Perception Attention Module (EPA) is employed to counteract information degradation due to low light conditions or partial obstructions, thus enhancing the texture details. The transformative process culminates with a Transformer module that integrates global contextual information and refines the final image output.

We summarize our contributions as follows:
\begin{enumerate}
    \item We propose an end-to-end Transformer model that not only achieves state-of-the-art performance on Visible-to-Infrared tasks but also minimizes computational overhead.
    \item We introduce the Dynamic Fusion Aggregation Module designed to integrate features extracted from visible light and map these features onto a latent space. This module enables a more precise capture and characterization of imagery information across diverse environments and conditions.
    \item We propose the Enhanced Perception Attention Module that mitigates information loss due to obstructions or low-light conditions, simultaneously enhancing the image's details and structure, thus augmenting the textural detail features of the image.
\end{enumerate}

\section{Related Work}

\subsection{Image-to-Image Translation}
Pix2Pix~\cite{isola2017image} introduces the concept of conditional GAN framework for image-to-image translation tasks. This supervised model requires paired images for training and demonstrates its effectiveness in scenarios where the input-output mapping is well-defined.

CycleGAN~\cite{zhu2017unpaired} extends the concept of image translation to unpaired datasets. By integrating a cycle consistency loss, CycleGAN addresses the limitation of paired training samples present in the Pix2Pix architecture, thus broadening the applicability to a wider range of domains such as photograph enhancement, art style transfer, and domain adaptation.

Building on the idea of unsupervised learning, Liu \etal proposed UNIT~\cite{liu2017unsupervised}, which assumes a shared latent space between different domains. Based on the shared-latent space assumption, UNIT incorporates Variational Autoencoders (VAEs) along with GANs to model the distribution of images in a joint framework.

Extending the capabilities of these models, MUNIT~\cite{huang2018multimodal} introduces a framework that addresses multimodal image translation. MUNIT disentangles the representation of content and style, allowing for the generation of diverse outputs from a single input image by manipulating the style code.

\subsection{Visible-to-Infrared Translation}
ThermalGAN~\cite{kniaz2018thermalgan} is a framework that utilizes generative adversarial networks (GANs) for cross-modality color-to-thermal image translation, specifically for person re-identification in multispectral datasets. It aims to generate realistic and diverse thermal images from color probe images, enabling effective matching and re-identification of individuals across different modalities. The ThermalGAN framework addresses the challenges of cross-modality matching by leveraging GANs to synthesize multimodal thermal probe sets from single color probe images, ultimately delivering robust matching performance that surpasses existing state-of-the-art methods in cross-modality color-thermal re-identification. This innovative approach holds significant potential for applications in multispectral imaging and person re-identification tasks. 

InfraGAN~\cite{ozkanouglu2022infragan} is a GAN architecture designed to generate the infrared (IR) equivalent of a given visible image. It aims to facilitate the use of multi-sensor-based applications by efficiently transferring visible images to the IR domain. InfraGAN utilizes a dual-generator architecture, where one generator generates the infrared images and the other generates the visible images. The model is trained in an adversarial manner to ensure the generated images are visually similar to real infrared images. Besides, InfraGAN incorporates a structural similarity loss function to focus on learning certain structural similarities between the IR and visible domains, while pixel-based $L_{1}$ norm enforces the architecture to look like an IR image. 

BCI~\cite{liu2022bci} is a pyramid pix2pix image generation method. This method uses a multi-scale approach to improve the quality of the generated images. It is based on the pix2pix algorithm, which uses a conditional generative adversarial network (cGAN) to learn a mapping between input and output images. By incorporating a multi-scale pyramid structure, the model can capture both global and local features of the input images and optimize the loss function at different scales, resulting in better image generation quality.

ClawGAN~\cite{luo2022clawgan} is an innovative approach to translating facial images from thermal to RGB visible light. It utilizes GAN and introduces claw connections to enhance the quality of infrared images. The framework is designed to overcome the difficulties of translating facial features from thermal to visible light, with a focus on improving the visual information and observability of image translation results in both bright and dark light conditions. Additionly, the framework integrates a mismatch metric for assessing the mapping relationship of paired images, employs template matching to minimize the mismatch, and incorporates synthesized and generative reconstructed loss functions to enhance the precision and quality of the image translation process.

\begin{figure*}[!tbh]
    \centering
    \includegraphics[width=1.0\textwidth]{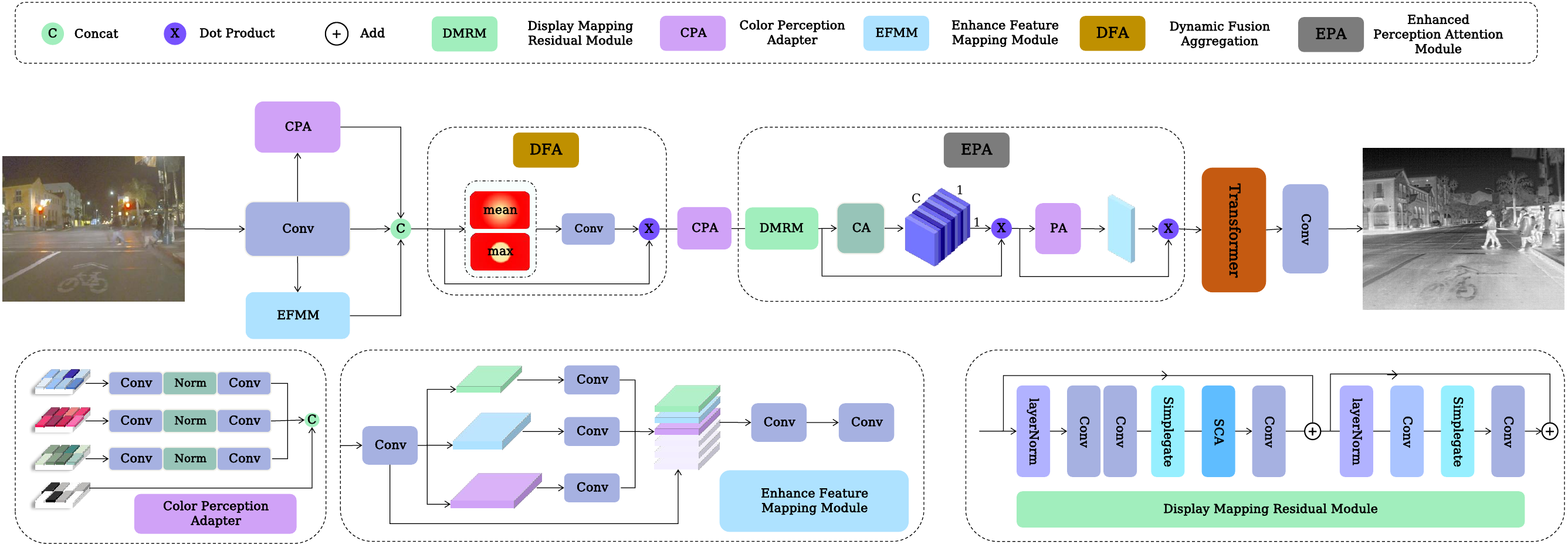}
    \caption{The overall architecture of the model. The image is processed through three parallel modules: Convolution (Conv), Color Perception Adapter (CPA), and Enhance Feature Mapping Module (EFM), which are responsible for extracting detail, color, and general convolutional features, respectively. These features are then amalgamated into a latent representation using the Dynamic Fusion Aggregation Module. Further, the CPA and the Enhanced Perception Attention mechanism act to transform these features to closely resemble those of an infrared image. Finally, a transformer module integrates global contextual information, serving to refine the final image output.}
    \label{fig:framework}
\end{figure*}

\section{Methodology}

\subsection{Overview}
We introduce a novel end-to-end model that fuses CNNs and Transformer architectures. The architecture of IRFormer is detailed in Figure~\ref{fig:framework}. Our model effectively encodes visible light information into a latent space and then decodes it into the infrared spectrum, placing particular focus on the retention of the intricate details unique to the visible light domain.

\subsection{Color Perception Adapter}
Conventional cameras capture visible light images comprising RGB color information, representing the red, green, and blue channels. These channels are attuned to the visible spectrum of light and provide critical information about the content within images, such as objects and scenes. Yet, they fall short in sensing the infrared spectrum, which can convey additional, non-visible information that may be crucial for certain applications. In contrast, infrared images primarily capture the infrared spectrum, unveiling distinctive details and characteristics of subjects that remain concealed in the visible spectrum.

To bridge this gap, our model integrates infrared spectral details by incorporating a novel component: a pair of color perception adapters (CPAs). These modules are pivotal for the translation of color features between the visible and infrared domains.

The first CPA is tasked with the extraction of infrared color features from visible light images—features that are present but imperceptible to the human eye. The subsequent CPA module takes these features and transposes them onto the infrared pixel domain. Through this process, the module learns to adapt to these features, enhancing the model's ability to interpret and reconstruct the infrared aspects of the input.

The CPA operates by segregating the input features into four distinct segments: the three RGB channels of visible light, and the infrared component. The RGB channels undergo a convolutional feature extraction process, which are then meticulously blended with the infrared features. The culmination of this synthesis is a unified representation that encompasses both visible and infrared spectral information. The result is a more accurate and comprehensive infrared representation generated from visible light inputs, enabling the model to produce outputs that are richer in detail and more useful for a variety of applications where infrared data is advantageous.

\subsection{Enhanced Feature Mapping Module}

Our model incorporates an Enhanced Feature Mapping Module specifically engineered to extract fine-grained texture features. This is achieved by employing downsampling at various scales to garner multi-scale feature representations~\cite{Li_2023_ICCV,ijcai2023p129,luo2023devignet}. The process of downsampling serves to decrease the spatial resolution of the input features, which in turn allows the model to discern broader-scale patterns and structures. Subsequently, these multi-scale features are processed through a sequence of convolutional layers and upsampling operations. This procedure culminates in the creation of multiple single-channel Detail Perception Enhancement Modules, each meticulously calibrated to match the dimensions of the initial input features. Once these enhancement modules have been established, they are incorporated with the original features and undergo additional convolutional processing. The outcome of this intricate process is a set of refined feature representations, distinguished by their enhanced detail and granularity. Such enriched features are inherently more discriminative and informative, thereby providing substantial benefits for subsequent tasks such as object detection and image classification.

\subsection{Dynamic Fusion Aggregation Module}
For the task of translating visible light images into the infrared spectrum, we focus on deriving all necessary features from the visible light data. This process is not only pivotal for teaching the model to understand the correlation between visible and infrared light but also for ensuring that the synthesized infrared images retain the texture characteristics of the original visible light images. To fulfill this requirement, we have engineered the Dynamic Fusion Aggregation Module (DFA), which is responsible for fusing features from visible light and projecting them into a latent space.This module is meticulously designed to ensure a seamless transition of visual information, preserving the integrity of the original features while preparing them for accurate infrared representation.

In this latent space, to emphasize the most salient features, we employ an attention mechanism. We construct an attention matrix that is informed by the maximum and average values of the feature sets. This matrix is designed to spotlight key features that are crucial for the translation process.

The subsequent feature mapping is guided by this attention-driven mechanism. The detailed formulation of this process is articulated in the following manner:
\begin{equation}
    f = [MaxPoold_{0}(x); AvgPoold_{0}(x)]
\end{equation}
\begin{equation}
    y = x\times \sigma (f(x))
\end{equation}
where $d_{0}$ is the 0-th dimension across which the max and average pooling operations take place. For instance, the $f$ of a tensor of shape $(C \times H \times W)$ results in a tensor of shape $(2 \times H \times W)$. The $\sigma$ represents the logistic function.

\subsection{Enhanced Perception Attention Module}
In tandem with the DFA, the Enhanced Perception Attention Module (EPA) plays a pivotal role in the model's translation process. It focuses on compensating for information loss that typically arises in low-light environments or areas with obstructed views. The EPA module utilizes a dual attention system, consisting of channel and pixel attention strategies, to meticulously refine and enhance the latent space's feature set. This dual mechanism ensures that each pixel is attentively adjusted to preserve and highlight critical textural details, culminating in infrared images that boast superior clarity and information content.

Inspired by NAFNet~\cite{chen2022simple}, we propose the Enhanced Perception Attention Module to select features for transforming infrared images from latent space to generate higher-quality infrared images. This module is designed to enhance feature representation to compensate for information loss caused by low lighting conditions or occlusions. The formulation of this process is as follows:
\begin{equation}
    Y = x + Conv(X_{1} \odot X_{2}) \times W_{1} 
\end{equation}
$Conv$ denotes the convolution operation, while $X_{1}$ and $X_{2}$ are two identical feature maps obtained by splitting the channels of $X$. $W_{1}$ represents a set of custom learnable weights.

Following the NAFNet architecture, we also utilize channel and pixel attention mechanisms to learn and compensate for the residual information that is often lost during challenging imaging scenarios. 

The channel attention mechanism (CA) is capable of capturing global information and is computationally efficient. It first compresses spatial information into the channel dimension and then applies a multi-layer perceptron to compute channel attention, which is used to weigh the feature maps. The pixel attention mechanism (PA) can adaptively adjusts the weights of each pixel, enabling the model to better focus on important pixels, thereby capturing subtle variations and details in the image, and extracting more informative features. The formulation of this process is as follows:
\begin{equation}
\begin{gathered}
CA(X) = X \times \sigma (W_{2}\max_{}(0,W_{1}pool(X)))\\
PA(X) = X \times \sigma (W_{4}\max_{}(0,W_{3}(X)))
\end{gathered}
\end{equation}
$\sigma$ is a nonlinear activation function, Sigmoid, $W_{1}$, $W_{2}$, $W_{3}$, $W_{4}$ are fully-connected layers and ReLU is adopted between two fully-connected layers. The features from $CA$ are processed through pooling, reducing the feature size from $(C, H, W)$ to $(C, 1, 1)$. The final convolutional output features have a size of $(C, 1, 1)$. The features from $PA$ are processed through convolution to transform the $(C, H, W)$ features into $(1, H, W)$.

\subsection{Transformer}
Inspired by Vision Transformer~\cite{zamir2022restormer,chen2023shadocnet,li2022wavenhancer,Zuo2023BrainFN,chen2023medprompt}, our model leverages the Transformer architecture to enhance the integration of information extrapolated by preceding modules. The Transformer's inherent proficiency in capturing long-range dependencies between features is particularly beneficial for completing and refining the infrared image reconstruction process.

\begin{table*}[ht]
\caption{
Quantitative comparison of various image translation methods across a range of datasets.
}
\adjustbox{width=\textwidth}{%
\begin{tabular}{l|ll|ll|ll|ll|ll|l|l}
\hline
\multirow{2}{*}{Method} &
  \multicolumn{2}{c|}{LLVIP} &
  \multicolumn{2}{c|}{RoadScene} &
  \multicolumn{2}{c|}{M3FD} &
  \multicolumn{2}{c|}{FLIR} &
  \multicolumn{2}{c|}{MCubeS} &
  \multicolumn{1}{c|}{\multirow{2}{*}{MACs(G)$\downarrow$}} &
  \multicolumn{1}{c}{\multirow{2}{*}{Params(M)$\downarrow$}} \\ \cline{2-11}
 &
  \multicolumn{1}{c}{PSNR$\uparrow$} &
  \multicolumn{1}{c|}{SSIM$\uparrow$} &
  \multicolumn{1}{c}{PSNR$\uparrow$} &
  \multicolumn{1}{c|}{SSIM$\uparrow$} &
  \multicolumn{1}{c}{PSNR$\uparrow$} &
  \multicolumn{1}{c|}{SSIM$\uparrow$} &
  \multicolumn{1}{c}{PSNR$\uparrow$} &
  \multicolumn{1}{c|}{SSIM$\uparrow$} &
  \multicolumn{1}{c}{PSNR$\uparrow$} &
  \multicolumn{1}{c|}{SSIM$\uparrow$} &
  \multicolumn{1}{c|}{} &
  \multicolumn{1}{c}{}
  \\ \hline
CycleGAN &
  \multicolumn{1}{c}{4.21} &
  \multicolumn{1}{c|}{0.12} &
  \multicolumn{1}{c}{4.19} &
  \multicolumn{1}{c|}{0.01} &
  \multicolumn{1}{c}{4.87} &
  \multicolumn{1}{c|}{0.07} &
  \multicolumn{1}{c}{3.45} &
  \multicolumn{1}{c|}{0.01} &
  \multicolumn{1}{c}{-1.30} &
  \multicolumn{1}{c|}{-0.13} &
  \multicolumn{1}{c|}{14.03} &
  \multicolumn{1}{c}{1.41} \\
Pix2Pix &
  \multicolumn{1}{c}{1.63} &
  \multicolumn{1}{c|}{0.01} &
  \multicolumn{1}{c}{4.68} &
  \multicolumn{1}{c|}{0.02} &
  \multicolumn{1}{c}{-0.74} &
  \multicolumn{1}{c|}{-0.12} &
  \multicolumn{1}{c}{4.19} &
  \multicolumn{1}{c|}{0.05} &
  \multicolumn{1}{c}{-2.58} &
  \multicolumn{1}{c|}{-0.12} &
  \multicolumn{1}{c|}{18.15} &
  \multicolumn{1}{c}{5.72} \\
UNIT &
   \multicolumn{1}{c}{3.58} &
  \multicolumn{1}{c|}{0.08} &
  \multicolumn{1}{c}{4.22} &
  \multicolumn{1}{c|}{0.00} &
  \multicolumn{1}{c}{5.28} &
  \multicolumn{1}{c|}{0.07} &
  \multicolumn{1}{c}{3.11} &
  \multicolumn{1}{c|}{0.01} &
  \multicolumn{1}{c}{-0.76} &
  \multicolumn{1}{c|}{-0.13} &
  \multicolumn{1}{c|}{61.42} &
  \multicolumn{1}{c}{1.35} \\
MUNIT &
   \multicolumn{1}{c}{4.06} &
  \multicolumn{1}{c|}{0.12} &
  \multicolumn{1}{c}{4.70} &
  \multicolumn{1}{c|}{0.03} &
  \multicolumn{1}{c}{4.53} &
  \multicolumn{1}{c|}{0.06} &
  \multicolumn{1}{c}{3.64} &
  \multicolumn{1}{c|}{0.02} &
  \multicolumn{1}{c}{-0.53} &
  \multicolumn{1}{c|}{-0.14} &
  \multicolumn{1}{c|}{66.31} &
  \multicolumn{1}{c}{1.85} \\
ThermalGAN &
   \multicolumn{1}{c}{-35.90} &
  \multicolumn{1}{c|}{0.00} &
  \multicolumn{1}{c}{-36.11} &
  \multicolumn{1}{c|}{0.00} &
  \multicolumn{1}{c}{-35.43} &
  \multicolumn{1}{c|}{-2.97} &
  \multicolumn{1}{c}{-36.00} &
  \multicolumn{1}{c|}{-3.35} &
  \multicolumn{1}{c}{-32.93} &
  \multicolumn{1}{c|}{0.00} &
  \multicolumn{1}{c|}{18.15} &
  \multicolumn{1}{c}{5.72} \\
BCI &
   \multicolumn{1}{c}{6.84} &
  \multicolumn{1}{c|}{0.11} &
  \multicolumn{1}{c}{7.67} &
  \multicolumn{1}{c|}{0.09} &
  \multicolumn{1}{c}{7.31} &
  \multicolumn{1}{c|}{0.15} &
  \multicolumn{1}{c}{11.14} &
  \multicolumn{1}{c|}{0.21} &
  \multicolumn{1}{c}{-2.21} &
  \multicolumn{1}{c|}{-0.16} &
  \multicolumn{1}{c|}{18.15} &
  \multicolumn{1}{c}{14.15} \\
InfraGAN &
   \multicolumn{1}{c}{-39.52} &
  \multicolumn{1}{c|}{0.00} &
  \multicolumn{1}{c}{-40.26} &
  \multicolumn{1}{c|}{0.00} &
  \multicolumn{1}{c}{-40.43} &
  \multicolumn{1}{c|}{0.00} &
  \multicolumn{1}{c}{-40.66} &
  \multicolumn{1}{c|}{1.85} &
  \multicolumn{1}{c}{-42.38} &
  \multicolumn{1}{c|}{-3.47} &
  \multicolumn{1}{c|}{18.15} &
  \multicolumn{1}{c}{5.44} \\
ClawGAN &
   \multicolumn{1}{c}{4.25} &
  \multicolumn{1}{c|}{0.12} &
  \multicolumn{1}{c}{5.09} &
  \multicolumn{1}{c|}{0.01} &
  \multicolumn{1}{c}{7.20} &
  \multicolumn{1}{c|}{0.03} &
  \multicolumn{1}{c}{5.91} &
  \multicolumn{1}{c|}{-0.04} &
  \multicolumn{1}{c}{-1.71} &
  \multicolumn{1}{c|}{-0.18} &
  \multicolumn{1}{c|}{14.03} &
  \multicolumn{1}{c}{0.51} \\ \hline
Ours &
  \multicolumn{1}{c}{\textbf{12.66}} &
  \multicolumn{1}{c|}{\textbf{0.44}} &
  \multicolumn{1}{c}{\textbf{14.35}} &
  \multicolumn{1}{c|}{\textbf{0.45}} &
  \multicolumn{1}{c}{\textbf{14.36}} &
  \multicolumn{1}{c|}{\textbf{0.56}} &
  \multicolumn{1}{c}{\textbf{17.74}} &
  \multicolumn{1}{c|}{\textbf{0.48}} &
  \multicolumn{1}{c}{\textbf{8.06}} &
  \multicolumn{1}{c|}{\textbf{0.79}} &
  \multicolumn{1}{c|}{\textbf{2.41}} &
  \multicolumn{1}{c}{\textbf{0.04}} \\ \hline
\end{tabular}%
}
\label{tab:comp}
\end{table*}

\begin{figure*}[ht]
 \begin{minipage}[c]{1.0\linewidth}
        % \begin{minipage}[b]{0.05\linewidth}
        %     \centering
        %     \centerline{}
        % \end{minipage}
        \begin{minipage}[b]{0.13\linewidth}
            \centering
            \centerline{Vis}
        \end{minipage}
        \begin{minipage}[b]{0.13\linewidth}
            \centering
            \centerline{CycleGAN}
        \end{minipage}
        \begin{minipage}[b]{0.13\linewidth}
            \centering
            \centerline{MUNIT}
        \end{minipage}
        \begin{minipage}[b]{0.13\linewidth}
            \centering
            \centerline{ClawGAN}
        \end{minipage}
        \begin{minipage}[b]{0.13\linewidth}
            \centering
            \centerline{BCI}
        \end{minipage}
        \begin{minipage}[b]{0.13\linewidth}
            \centering
            \centerline{Ours}
        \end{minipage}
        \begin{minipage}[b]{0.13\linewidth}
            \centering
            \centerline{IR}
        \end{minipage}
    \end{minipage}
%first line
     \begin{minipage}[c]{1.0\linewidth}
     % \begin{minipage}[c]{0.05\linewidth}
     %        \centering
     %        \centerline{LLVIP}
     %    \end{minipage}
        \begin{minipage}[b]{0.13\linewidth}
            \centering
            \centerline{\includegraphics[width=\linewidth]{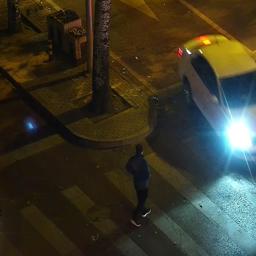}}
        \end{minipage}
        \begin{minipage}[b]{0.13\linewidth}
            \centering
            \centerline{\includegraphics[width=\linewidth]{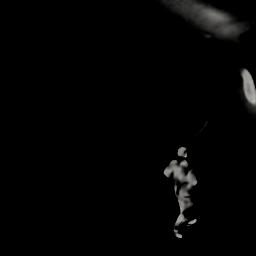}}  
        \end{minipage}
        \begin{minipage}[b]{0.13\linewidth}
            \centering
            \centerline{\includegraphics[width=\linewidth]{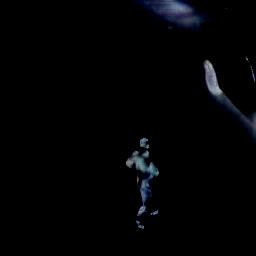}}
        \end{minipage}
        \begin{minipage}[b]{0.13\linewidth}
            \centering
            \centerline{\includegraphics[width=\linewidth]{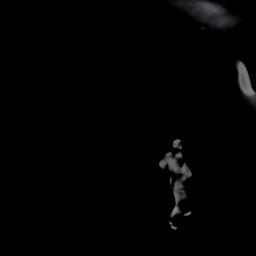}}
        \end{minipage}
        \begin{minipage}[b]{0.13\linewidth}
            \centering
            \centerline{\includegraphics[width=\linewidth]{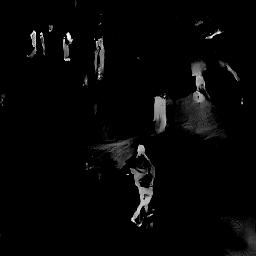}}
        \end{minipage}
        \begin{minipage}[b]{0.13\linewidth}
            \centering
            \centerline{\includegraphics[width=\linewidth]{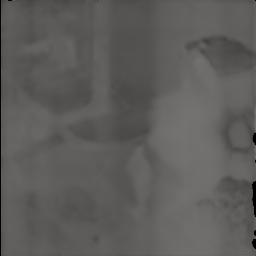}}
        \end{minipage}
        \begin{minipage}[b]{0.13\linewidth}
            \centering
            \centerline{\includegraphics[width=\linewidth]{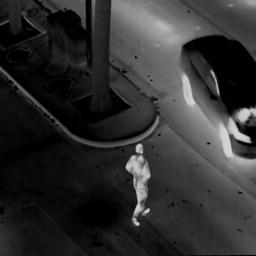}}
        \end{minipage}
    \end{minipage}
% second
    \begin{minipage}[c]{1.0\linewidth}
     % \begin{minipage}[c]{0.05\linewidth}
     %        \centering
     %        \centerline{RoadScene}
     %    \end{minipage}
        \begin{minipage}[b]{0.13\linewidth}
            \centering
            \centerline{\includegraphics[width=\linewidth]{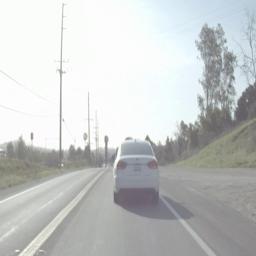}}
        \end{minipage}
        \begin{minipage}[b]{0.13\linewidth}
            \centering
            \centerline{\includegraphics[width=\linewidth]{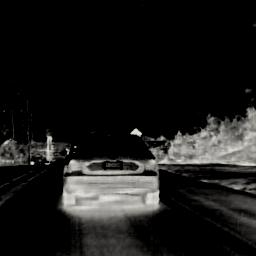}}  
        \end{minipage}
        \begin{minipage}[b]{0.13\linewidth}
            \centering
            \centerline{\includegraphics[width=\linewidth]{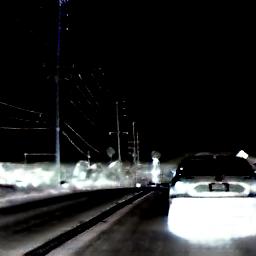}}
        \end{minipage}
        \begin{minipage}[b]{0.13\linewidth}
            \centering
            \centerline{\includegraphics[width=\linewidth]{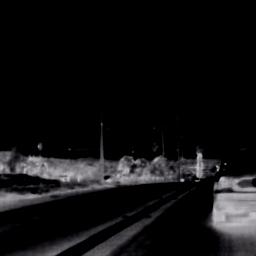}}
        \end{minipage}
        \begin{minipage}[b]{0.13\linewidth}
            \centering
            \centerline{\includegraphics[width=\linewidth]{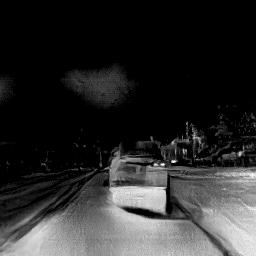}}
        \end{minipage}
        \begin{minipage}[b]{0.13\linewidth}
            \centering
            \centerline{\includegraphics[width=\linewidth]{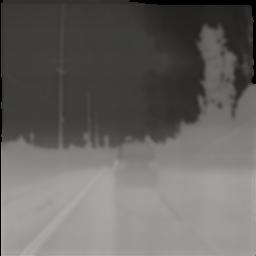}}
        \end{minipage}
        \begin{minipage}[b]{0.13\linewidth}
            \centering
            \centerline{\includegraphics[width=\linewidth]{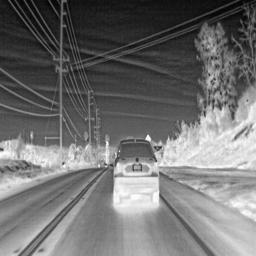}}
        \end{minipage}
    \end{minipage}
    \begin{minipage}[c]{1.0\linewidth}
     % \begin{minipage}[c]{0.05\linewidth}
     %        \centering
     %        \centerline{M3FD}
     %    \end{minipage}
        \begin{minipage}[b]{0.13\linewidth}
            \centering
            \centerline{\includegraphics[width=\linewidth]{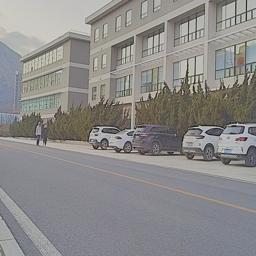}}
        \end{minipage}
        \begin{minipage}[b]{0.13\linewidth}
            \centering
            \centerline{\includegraphics[width=\linewidth]{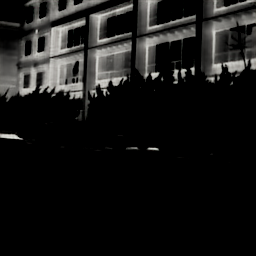}}  
        \end{minipage}
        \begin{minipage}[b]{0.13\linewidth}
            \centering
            \centerline{\includegraphics[width=\linewidth]{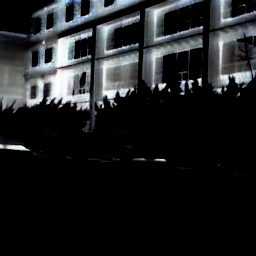}}
        \end{minipage}
        \begin{minipage}[b]{0.13\linewidth}
            \centering
            \centerline{\includegraphics[width=\linewidth]{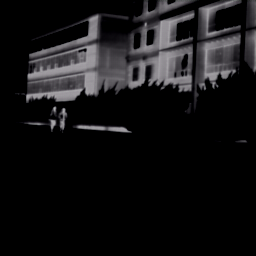}}
        \end{minipage}
        \begin{minipage}[b]{0.13\linewidth}
            \centering
            \centerline{\includegraphics[width=\linewidth]{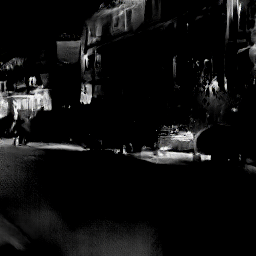}}
        \end{minipage}
        \begin{minipage}[b]{0.13\linewidth}
            \centering
            \centerline{\includegraphics[width=\linewidth]{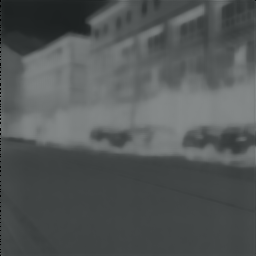}}
        \end{minipage}
        \begin{minipage}[b]{0.13\linewidth}
            \centering
            \centerline{\includegraphics[width=\linewidth]{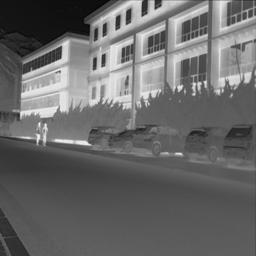}}
        \end{minipage}
    \end{minipage}
    \begin{minipage}[c]{1.0\linewidth}
     % \begin{minipage}[c]{0.05\linewidth}
     %        \centering
     %        \centerline{FLIR}
     %    \end{minipage}
        \begin{minipage}[b]{0.13\linewidth}
            \centering
            \centerline{\includegraphics[width=\linewidth]{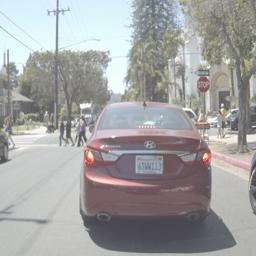}}
        \end{minipage}
        \begin{minipage}[b]{0.13\linewidth}
            \centering
            \centerline{\includegraphics[width=\linewidth]{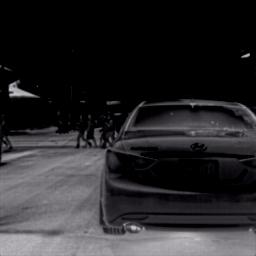}}  
        \end{minipage}
        \begin{minipage}[b]{0.13\linewidth}
            \centering
            \centerline{\includegraphics[width=\linewidth]{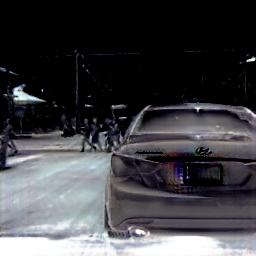}}
        \end{minipage}
        \begin{minipage}[b]{0.13\linewidth}
            \centering
            \centerline{\includegraphics[width=\linewidth]{figs/CGAN_FLIR.jpg}}
        \end{minipage}
        \begin{minipage}[b]{0.13\linewidth}
            \centering
            \centerline{\includegraphics[width=\linewidth]{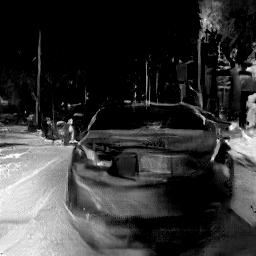}}
        \end{minipage}
        \begin{minipage}[b]{0.13\linewidth}
            \centering
            \centerline{\includegraphics[width=\linewidth]{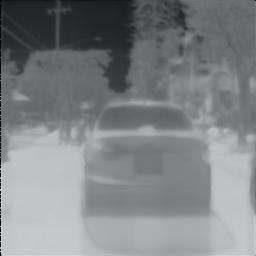}}
        \end{minipage}
        \begin{minipage}[b]{0.13\linewidth}
            \centering
            \centerline{\includegraphics[width=\linewidth]{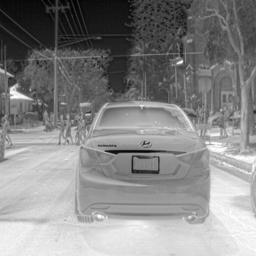}}
        \end{minipage}
    \end{minipage}
    \begin{minipage}[c]{1.0\linewidth}
     % \begin{minipage}[c]{0.05\linewidth}
     %        \centering
     %        \centerline{MCubeS}
     %    \end{minipage}
        \begin{minipage}[b]{0.13\linewidth}
            \centering
            \centerline{\includegraphics[width=\linewidth]{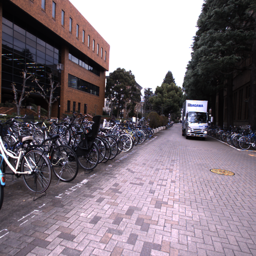}}
        \end{minipage}
        \begin{minipage}[b]{0.13\linewidth}
            \centering
            \centerline{\includegraphics[width=\linewidth]{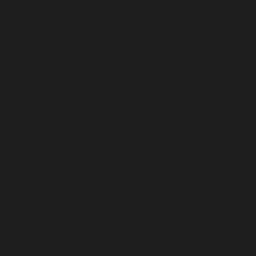}}  
        \end{minipage}
        \begin{minipage}[b]{0.13\linewidth}
            \centering
            \centerline{\includegraphics[width=\linewidth]{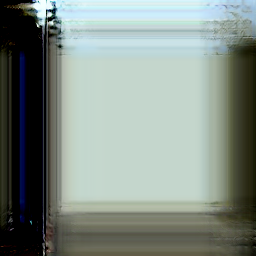}}
        \end{minipage}
        \begin{minipage}[b]{0.13\linewidth}
            \centering
            \centerline{\includegraphics[width=\linewidth]{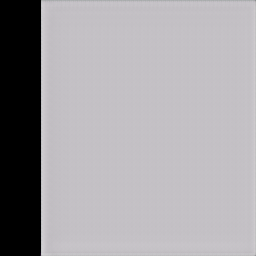}}
        \end{minipage}
        \begin{minipage}[b]{0.13\linewidth}
            \centering
            \centerline{\includegraphics[width=\linewidth]{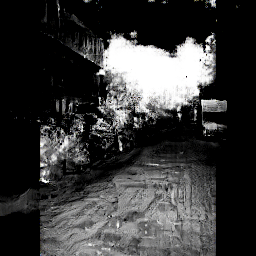}}
        \end{minipage}
        \begin{minipage}[b]{0.13\linewidth}
            \centering
            \centerline{\includegraphics[width=\linewidth]{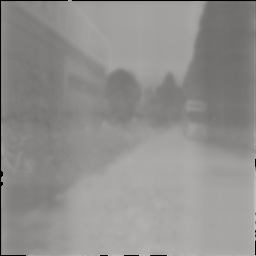}}
        \end{minipage}
        \begin{minipage}[b]{0.13\linewidth}
            \centering
            \centerline{\includegraphics[width=\linewidth]{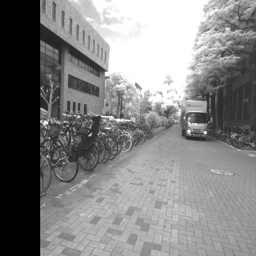}}
        \end{minipage}
    \end{minipage}
    \caption{The visual comparison across five datasets—LLVIP, RoadScene, M3FD, FLIR, and MCubeS—highlights the performance from top to bottom. }
    \label{fig:comp}
\end{figure*}

\begin{figure*}[t]
 \begin{minipage}[c]{1.0\linewidth}
        % \begin{minipage}[b]{0.05\linewidth}
        %     \centering
        %     \centerline{}
        % \end{minipage}
        \begin{minipage}[b]{0.16\linewidth}
            \centering
            \centerline{CycleGAN}
        \end{minipage}
        \begin{minipage}[b]{0.16\linewidth}
            \centering
            \centerline{MUNIT}
        \end{minipage}
        \begin{minipage}[b]{0.16\linewidth}
            \centering
            \centerline{ClawGAN}
        \end{minipage}
        \begin{minipage}[b]{0.16\linewidth}
            \centering
            \centerline{BCI}
        \end{minipage}
        \begin{minipage}[b]{0.16\linewidth}
            \centering
            \centerline{Ours}
        \end{minipage}
        \begin{minipage}[b]{0.16\linewidth}
            \centering
            \centerline{IR}
        \end{minipage}
    \end{minipage}
%first line
     \begin{minipage}[c]{1.0\linewidth}
     % \begin{minipage}[c]{0.05\linewidth}
     %        \centering
     %        \centerline{LLVIP}
     %    \end{minipage}
        \begin{minipage}[b]{0.16\linewidth}
            \centering
            \centerline{\includegraphics[width=\linewidth]{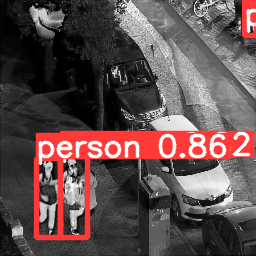}}  
        \end{minipage}
        \begin{minipage}[b]{0.16\linewidth}
            \centering
            \centerline{\includegraphics[width=\linewidth]{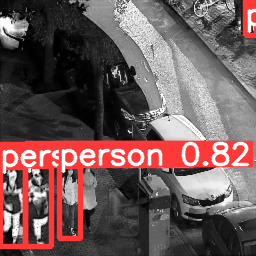}}
        \end{minipage}
        \begin{minipage}[b]{0.16\linewidth}
            \centering
            \centerline{\includegraphics[width=\linewidth]{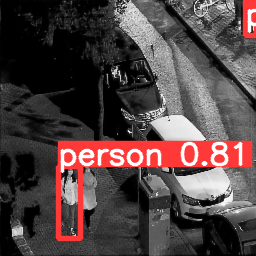}}
        \end{minipage}
        \begin{minipage}[b]{0.16\linewidth}
            \centering
            \centerline{\includegraphics[width=\linewidth]{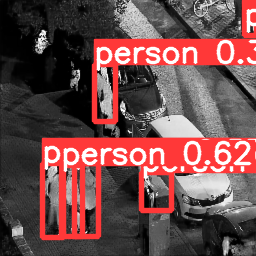}}
        \end{minipage}
        \begin{minipage}[b]{0.16\linewidth}
            \centering
            \centerline{\includegraphics[width=\linewidth]{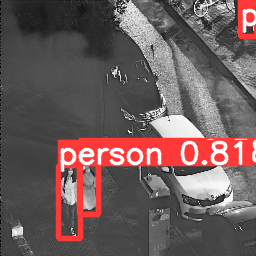}}
        \end{minipage}
        \begin{minipage}[b]{0.16\linewidth}
            \centering
            \centerline{\includegraphics[width=\linewidth]{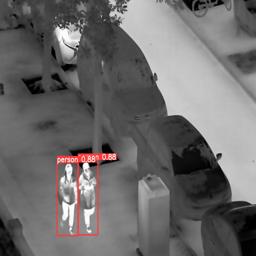}}
        \end{minipage}
    \end{minipage}
    \caption{
    In comparing detection performances for downstream applications, it has been noted that the infrared images generated by CycleGAN, MUNIT, ClawGAN, and BCI are frequently marred by the inclusion of irrelevant human silhouettes. The introduction of these artifacts can lead to undetected objects and a higher incidence of false detection.
    }
    \label{fig:det}
\end{figure*}

\subsection{Objective Function}
To better preserve structural information and enhance image contrast, our model adopts a dual loss function strategy. This strategy synergizes the benefits of the Smooth $L_{1}$ loss function, celebrated for its robustness to outliers and facilitation of pixel-level accuracy, with the Structural Similarity Index Measure $L_{SSIM}$. $L_{SSIM}$ is a metric devised to assess the perceptual quality of images by considering their luminance, contrast, and structural similarities.

The $L_{1}$ and $L_{SSIM}$ loss function is mathematically represented as follows:
\begin{equation}
\begin{gathered}
L_{smooth}(x)=\left\{\begin{matrix}
 0.5\times x^{2}  &  &  &\left | x \right |<1  \\
  \left | x \right |-0.5 &  &  & otherwise
\end{matrix}\right.
\end{gathered}
\end{equation}
\begin{equation}
\begin{gathered}
L_{SSIM}(x,y)=\frac{(2\mu _{x}\mu_{y} +C_{1} )(2\sigma _{xy}+C_{2})}{(\mu_{x}^{2}+\mu_{y}^{2}+C_{1} )(\sigma_{x}^{2}+\sigma_{y}^{2}+C_{2})} 
\end{gathered}
\end{equation}
$\mu_{x}$ and $\mu_{y}$ represent the mean values (averages) of $x$ and $y$, respectively.$\sigma_{x}^{2}$ and $\sigma_{y}^{2}$ are the variances of $x$ and $y$, respectively. $\sigma_{xy}$ is the covariance between $x$ and $y$. $C_{1}=(k_{1}L)^{2}$ and $C_{2}=(k_{2}L)^{2}$ are two constants used to stabilize division when the denominator approaches zero. Here, $L$ represents the dynamic range of the image data, and $k_{1}$ and $k_{2}$ are two very small constants.

Our total loss function is represented as:
\begin{equation}
L(x,y)=L_{smooth}(x)+L_{SSIM}(x,y) 
\end{equation}
In this context, $L$ denotes the total loss function, $L_{smooth}$ represents the smooth $L_{1}$ loss component, and $L_{SSIM}$ signifies the loss component calculated based on the Structural Similarity Index Measure.

\section{Experiments}

\subsection{Datasets and Metrics}

In order to improve the robustness and precision of our model's training process, we combine training sets from five separate datasets into one comprehensive training regimen. This integration aims to expand the diversity of the data and improve the model's generalizability.

\subsubsection{LLVIP~\cite{jia2021llvip}} This dataset consists of 15,488 pairs of images and is primarily designed for extremely low-light circumstances. Precise temporal and spatial alignment of visible and infrared images is a distinguishing feature of LLVIP, making it exceptionally useful for pedestrian detection tasks due to its meticulously labeled pedestrian data.

\subsubsection{RoadScene~\cite{xu2020fusiondn}} With 221 aligned pairs of visible and infrared images, RoadScene offers a varied compilation of scenarios including roads, vehicles, and pedestrians. It serves as a resource for road scene analysis and is applicable to tasks like feature matching, image registration, and image fusion.

\subsubsection{M3FD~\cite{liu2022target}} This dataset encompasses 8,400 images for tasks such as fusion, detection, and fusion-based detection, along with 600 images specifically for scene fusion. Target labels in M3FD include six categories: people, cars, buses, motorcycles, lamps, and trucks, providing extensive ground truth for various detection algorithms.

\subsubsection{FLIR} Dedicated to promoting advancements in fusion algorithms for visible and thermal sensor data, the FLIR dataset presents unaligned visible and infrared images. We have employed an automated alignment algorithm to accurately pair 5,142 sets of images, thereby making the dataset suitable for our training purposes.

\subsubsection{MCubeS~\cite{Liang_2022_CVPR}} This multimodal material segmentation dataset features images obtained from identical viewpoints but with different imaging modalities, including RGB, polarization, and near-infrared. It contains a total of 20 material categories, each meticulously labeled on a per-pixel basis for material category recognition.

For a more objective evaluation of our model's performance, Peak Signal-to-Noise Ratio (PSNR) and Structural Similarity Index Measure (SSIM) are utilized as assessment metrics. PSNR determines image quality by calculating the Mean Squared Error (MSE) between the original image and its distorted counterpart, such as a compressed version. SSIM, in contrast, measures the similarity between two images by analyzing elements of luminance, contrast, and structure. Higher values of PSNR and SSIM indicate superior visual quality and greater structural resemblance, respectively.

\subsection{Implementation Details} 
Our model utilizes the PyTorch neural network framework to process input images in both the visible and infrared spectra, with each image having a resolution of $256 \times 256$ pixels. To fine-tune the network's hyperparameters, we begin with an initial learning rate of $2 \times 10^{-4}$ and employ the AdamW~\cite{Loshchilov2017DecoupledWD} optimizer, which is known for its efficient handling of weight decay.

To dynamically adjust the learning rate throughout the training process, the CosineAnnealingLR~\cite{Loshchilov2016SGDRSG,li2023optimized} scheduler is utilized. This scheduler reduces the learning rate according to a cosine annealing schedule, allowing for a more refined convergence of the model by modulating the learning rate in a periodic yet diminishing fashion.

The training is conducted over 400 epochs to ensure thorough learning and convergence of the model.  We apply a batch size of 1, which may increase computation time, but it leads to more precise gradient calculations, leading to improved learning efficacy and model stability.

\subsection{Comparisons with State-of-the-Art Methods}
In order to evaluate the effectiveness of our method, we conducted comparison experiments using the aforementioned five datasets. Our method has been benchmarked against a suite of infrared image generation algorithms: CycleGAN, Pix2Pix, UNIT, MUNIT, ThermalGAN, BCI, InfraGAN, and ClawGAN. For standardization, we resized all input images to $256 \times 256$ pixels for processing by these models.

Table~\ref{tab:comp} presents an extensive and objective assessment of the experimental results. This table collects and quantifies the performance metrics for each algorithm, thereby facilitating a more comprehensive and quantitative analysis of their effectiveness. Our model distinguishes itself through its novel architecture and sophisticated integration of modules, each contributing to the superior quality of the translated infrared images. The Dynamic Fusion Aggregation Module (DFA) plays an essential role in integrating features extracted from the visible spectrum and projecting them into a latent space that mediates between visible and infrared domains. The design of this module enables a more precise and contextual representation of the imagery, which is crucial when dealing with the inherent variability of environmental conditions and lighting in various settings. The Enhanced Perception Attention Module (EPA) significantly contributes to the model's effectiveness by mitigating information loss due to occlusions or low-light conditions. This module enhances the image's details and structure, ensuring that textural detail features are pronounced and preserved in the translated infrared images. It is crucial to preserve the accuracy of features that are frequently degraded in situations with low contrast. Additionally, our model achieves these results with remarkable efficiency, maintaining a low computational overhead with only 0.04M parameters and 2.41G Multiply-Accumulate operations (MACs). This efficiency makes it a practical and scalable solution for real-world applications, where the demand for high-quality infrared imaging is balanced with the need for computational resourcefulness.

Figure~\ref{fig:comp} displays a subjective visual comparison of select outputs from these experiments. This figure visually demonstrates the performance of each algorithm for generating infrared images, serving as a benchmark to showcase their capabilities.

\subsection{Downstream Applications}
To prove the increased applicability of our generated IR images for downstream applications, we utilize the LLVIP dataset and integrate visible light images with our output IR images using the CDDFuse model~\cite{zhao2023cddfuse}. Following this fusion, we conduct pedestrian detection tasks with the YOLOv5~\cite{ultralytics2021yolov5}. The evaluation metric employed to gauge object detection performance is mean Average Precision (mAP). We calculated the mAP values at two confidence thresholds, specifically 50\% and 95\%, to precisely quantify the performance in this downstream task. The outcomes are recorded in Table~\ref{tab:det}, which outlines the performance metrics attained in the pedestrian detection task. This table provides a quantitative evaluation, illustrating the effectiveness of our generated images in improving the precision and dependability of pedestrian detection.

\begin{table}[ht]
\caption{
Ablation study on different modules of the proposed model.
}
\centering
{
\begin{tabular}{ccccc}
\hline
\multirow{2}{*}{Index} & \multirow{2}{*}{EPA} & \multirow{2}{*}{DFA}&\multicolumn{2}{c}{Metric} \\ 
\cline{4-5} & & & PSNR $\uparrow$ &SSIM$\uparrow$
\\ \hline
(1) & $\surd$ & &13.85 &0.47\\
(2) &  &$\surd$&13.96&0.47\\
(3) & $\surd$ &$\surd$&\textbf{14.01}&\textbf{0.48}\\
\hline
\end{tabular}
}
\label{tab:ab}
\end{table}

\begin{table}[ht]
\centering
\caption{
The comparative performance of different models on a pedestrian detection downstream task.
}
\begin{tabular}{l|c|c}
\hline
Method     & mAP\_50$\uparrow$ & mAP\_95$\uparrow$ \\ \hline
CycleGAN   & 0.133   & 0.042   \\
Pix2Pix    & 0.145   & 0.054   \\
UNIT       & 0.114   & 0.037   \\
MUNIT      & 0.136   & 0.042   \\
ThermalGAN & 0.133   & 0.047   \\
BCI        & 0.114   & 0.037   \\
InfraGAN   & 0.135   & 0.047   \\
ClawGAN    & 0.124   & 0.050   \\ \hline
Ours       & \textbf{0.146}   & \textbf{0.056}   \\ \hline
\end{tabular}
\label{tab:det}
\end{table}

\subsection{Ablation Studies}
To evaluate the effectiveness of our proposed modules, we conducted ablation experiments using a combined test set that included samples from all the datasets. The first experiment entails removing the Enhanced Perception Attention Module (EPA), while leaving the other modules intact. In the second experiment, we substitute the Dynamic Fusion Aggregation Module (DFA) with a convolutional module, ensuring the number of channels remained the same. The results of these experiments are detailed in Table~\ref{tab:ab}.

\section{Conclusion}
In this study, we propose a Transformer-based model that efficiently translates visible light images into high-fidelity infrared images. The proposed Dynamic Fusion Aggregation Module and Enhanced Perception Attention Module synergistically work together to effectively capture and retain crucial textural and color features, which are then accurately converted to the infrared domain. The effectiveness of our model is demonstrated through extensive benchmarking experiments, which show a clear superiority over existing methods in producing qualitatively and quantitatively superior infrared images. Moreover, the architecture of our model guarantees an appropriate trade-off between the efficiency of feature extraction and the computing burden, making it an appealing choice for real-life scenarios that demand superior infrared imaging. Future work may focus on further enhancing image contrast and overcoming any remaining limitations in contrast translation to fully exploit the distinct characteristics of infrared imagery.

\section*{Acknowledgment}
This work was supported in part by the Guangdong Provincial Key R\&D Programme under Grant No.2023B1111050010 and No.2020B0101100001, in part by the Huizhou Daya Bay Science and Technology Planning Project under Grant No.2020020003.

\bibliographystyle{IEEEtran}
\bibliography{ref}

\end{document}